\documentclass{article}
\usepackage{spconf,amsmath,graphicx, multirow}
\usepackage{float}
\usepackage[hyphens]{url}

\title{Region Ensemble Network: Improving Convolutional Network for Hand Pose Estimation}

\name{Hengkai Guo, Guijin Wang, Xinghao Chen, Cairong Zhang, Fei Qiao, Huazhong Yang}
\address{Department of Electronic Engineering\\
	Tsinghua University, Beijing}

\begin{document}
%\ninept
%
\maketitle
\begin{abstract}
Hand pose estimation from monocular depth images is an important and challenging problem for human-computer interaction. Recently deep convolutional networks (ConvNet) with sophisticated design have been employed to address it, but the improvement over traditional methods is not so apparent. To promote the performance of directly 3D coordinate regression, we propose a tree-structured Region Ensemble Network (REN), which partitions the convolution outputs into regions and integrates the results from multiple regressors on each regions. Compared with multi-model ensemble, our model is completely end-to-end training. The experimental results demonstrate that our approach achieves the best performance among state-of-the-arts on two public datasets.
\end{abstract}
\section{Introduction}
Hand pose estimation from single depth image plays an important role in applications of human-computer interface (HCI) and augmented reality (AR). Though has been studied for several years \cite{supancic2015depth}, it is still challenging due to large view variance, high joint flexibility, poor depth quality, severe self occlusion and similar part confusion.

Recently, convolutional networks (ConvNets) have witnessed great growth in several computer vision tasks such as object classification \cite{krizhevsky2012imagenet} and human pose estimation \cite{bulat2016human} because of great modeling capacity and end-to-end feature learning. ConvNets have also been introduced to solve the problem of hand pose estimation, often with complicated structure design such as multi-branch inputs \cite{tompson2014real}\cite{oberweger2015hands} and multi-model regression \cite{oberweger2015hands} \cite{oberwegertraining} \cite{gerobust} \cite{zhang2016learning}. However, ConvNets remain unable to obtain significant advantage over traditional random forest based methods \cite{sun2015cascaded} \cite{wan2016hand}. 

\begin{figure}[htb]
\centering
{\includegraphics[width=0.5\textwidth]{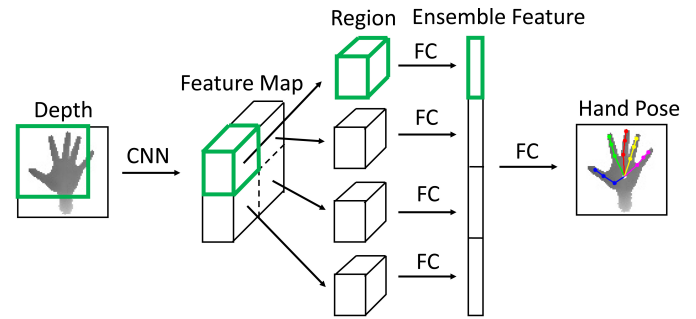}}
\caption{Overview of region ensemble network (REN).}
\label{fig_overview}
\end{figure}

Inspired by model ensemble and multi-view voting \cite{krizhevsky2012imagenet}, we present a single ConvNet architecture named \emph{Region Ensemble Net (REN)} \footnote{Codes are available at \url{https://github.com/guohengkai/region-ensemble-network}} (Fig.\ref{fig_overview}) to directly regress the 3D joint coordinates in monocular depth images with end-to-end optimization and inference. We implement it by training individual fully-connected (FC) layers on multiple feature regions and combining them as ensembles. As shown in our experiments, REN significantly promotes the performance of our ConvNet, which outperforms all state-of-the-art methods on two challenging hand pose benchmarks \cite{tompson2014real} \cite{tang2014latent}.

\section{Related work}
\noindent\textbf{Hand pose estimation with ConvNets}\hspace{2mm} 
Recently deep ConvNets have been applied on hand pose estimation for depth imaging \cite{wang2013depth}\cite{shi2015high}. Tompson et al. \cite{tompson2014real} use ConvNets to produce 2D heat maps and infer the 3D hand pose with inverse kinematics. Oberweger et al. \cite{oberweger2015hands} directly regress the 3D positions with multi-stage ConvNets using a linear layer as pose prior. In \cite{oberwegertraining}, a feedback loop is employed to iteratively correct the mistake, in which 3 ConvNets are used for initialization, synthesis and pose updating. Ge et al. \cite{gerobust} employ 3 ConvNets to separately regress 2D heat maps for each view with depth projections and fuse them to produce 3D hand pose. In \cite{zhou2016model}, physical joint constraints are incorporated into a forward kinematics based layer in ConvNet. Similarly, Zhang et al. \cite{zhang2016learning} embeds skeletal manifold into ConvNets and trains the model end-to-end to render sequential prediction.

\noindent\textbf{Multi-model ensemble methods for ConvNets}\hspace{2mm}
Traditional ensemble learning means that training multiple individual models and combining their outputs via averaging or weighted fusions, which is widely adopted in recognition competitions \cite{krizhevsky2012imagenet}. In addition to bagging \cite{krizhevsky2012imagenet} \cite{sun2013deep}, boosting is also introduced for people counting \cite{walach2016learning}. However, using multiple ConvNets requires large memory and time, which is not practical for applications.

\noindent\textbf{Multi-branch ensemble methods for ConvNets}\hspace{2mm}
We view single ConvNet with the fusion of multiple branches as a generalized type of ensemble. One popular strategy is to fuse different scaling inputs \cite{tompson2014real} \cite{oberweger2015hands} or different image cues \cite{guo2016two} \cite{li2016deeptrack} \cite{chen2016accurate} with multi-input branches. Another approach is to employ multi-output branches with shared convolutional feature extractor, either training with different samples \cite{li2016convolutional} or learning to predict different categories \cite{ahmed2016network}. Compared with multi-input, multi-output methods cost less time because inference of FC layers is much faster than that of convolutional layers. Our method also falls into such category.

\noindent\textbf{Multi-view testing for ConvNets}\hspace{2mm}
Multi-view testing is widely used to improve accuracy for object classification \cite{krizhevsky2012imagenet}. In \cite{krizhevsky2012imagenet}, predictions from 10-crop (4 corner and 1 center with horizontal flip) are averaged on single ConvNet. In \cite{sermanet2013overfeat}, fully-convolutional networks are employed in testing with inputs of multi-scale and multi-view and average pooling is applied on the score map to obtain the final scores. Such strategy has not been applied on hand pose estimation yet.

\section{Region Ensemble Network (REN)}
As in Fig.\ref{fig_overview}, REN starts with a ConvNet for feature extraction. Then the features are divided into multiple grid regions. Each region is fed into FC layers and learnt to fuse for pose prediction. In this section we introduce the basic network architecture, region ensemble structure and implementation details.

\begin{figure}[htb]
\centering
{\includegraphics[width=0.5\textwidth]{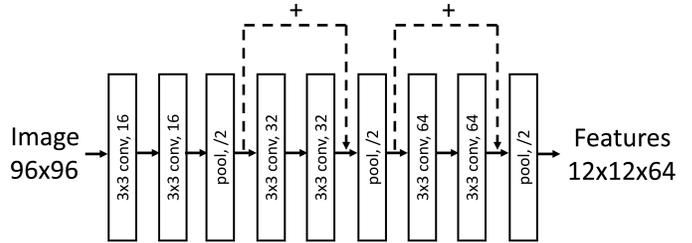}}
\caption{Structure of ConvNet for feature extraction. The dotted arrows represent residual connections \cite{he2015deep}.}
\label{fig_baseline}
\end{figure}

\noindent\textbf{Network architecture}\hspace{2mm} The architecture of our ConvNet for feature extraction consists of six $3\times3$ convolution layers (Fig.\ref{fig_baseline}) that accepts a $96\times96$ depth image as inputs. Each convolution layer is followed by a Rectified Linear Unit (ReLU) activation. Two residual connections \cite{he2015deep} are adopted between pooling layers with $1\times1$ convolution filters for dimension increase. The dimension of output feature maps is $12\times12\times64$. For regression, we use two 2048 dimension FC layers with dropout rate of 0.5 for each regressor. The output of regressor is a $3\times J$ vector representing the 3D world coordinates for hand joints, where $J$ is the number of joints.

\noindent\textbf{Region Ensemble}\hspace{2mm} Multi-view testing averages predictions from different crops of original image, which reduces the varaince for image classification \cite{krizhevsky2012imagenet}. Directly using multiple inputs is time-consuming. Because each activation in the convolutional feature maps is contributed by a receptive field in the image domain, we can project the multi-view inputs onto the regions of the feature maps. So multi-view voting is equal to utilizing each regions to separately predict the whole hand pose and combining the results. Based on it, we define a tree-structured network consisting of a single ConvNet trunk and several regression branches (Fig.\ref{fig_overview}). We uniformly divide the feature maps of ConvNet into an $n \times n$ grid. For each grid region, we feed it into the FC layers respectively as branches. A simple strategy for combination of different branches is bagging, which averages all outputs of branches. To better boost the predictions from all the regions, we employ region ensemble strategy instead of bagging: features from the last FC layers of all regions are concatenated and used to infer the hand pose with an extra regression layer. The whole network can be trained end-to-end by minimizing the regression loss. We set $n = 2$ to balance the trade-off between performance and efficiency, so the receptive field of single region within the $96\times96$ image bounding is $62\times62$, which can be seen as the corner crop in \cite{krizhevsky2012imagenet}. Including the center crop does not provide any further increase in accuracy. Note that we do not adopt multi-scale regions because it will lead to imbalanced parameter number in FC layers.

There are three main differences between proposed methods and multi-view voting: 1) To our knowledge, all multi-view testing methods before are designed for image classification while region ensemble can be applied on both classification and regression. 2) We adopt end-to-end training for region ensemble instead of testing only, making the ConvNet adjust the contributions from each views. 3) We replace the average pooling with FC on concatenated features to learn to fuse, which increases the learning ability of the network.

\noindent\textbf{Implementation}\hspace{2mm} We follow the practice in \cite{tompson2014real} \cite{oberweger2015hands} using Caffe \cite{jia2014caffe}. We first segment the foreground and extract a cube of size 150mm from the depth image centered in the centroid of hand region. Then the cube is resized into $96\times96$ patch of depth values normalized to $[-1, 1]$ as input for ConvNet, with data augmentation of random translation, scaling and rotation. We use stochastic gradient descent (SGD) with a mini-batch size of 128. The learning rate starts from 0.005 and is divided by 10 after every 50000 iterations, and the model is trained for up to 200000 iterations. In the meanwhile, we use a weight decay of 0.0005 and a momentum of 0.9.

\section{Experiments}
\begin{figure*}[htb]
\centering
{\includegraphics[width=0.95\textwidth]{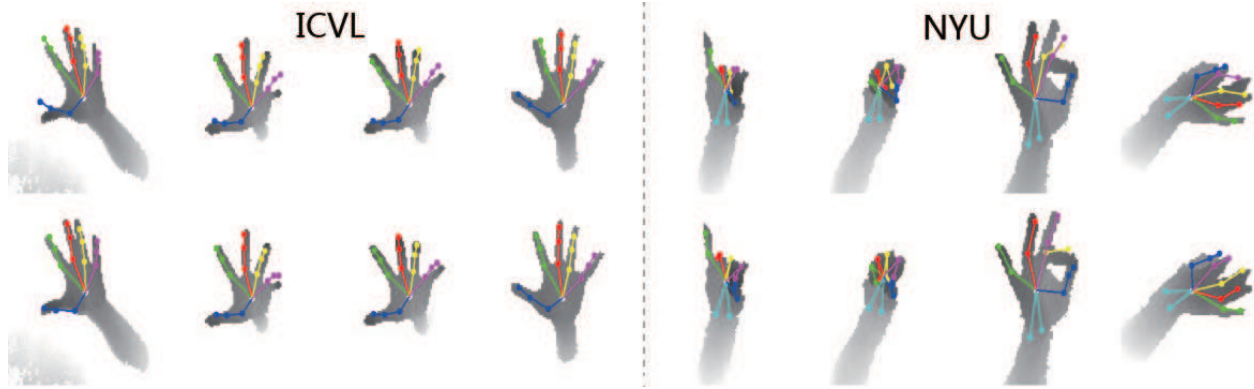}}
\caption{Example results on ICVL \cite{tang2014latent} and NYU \cite{tompson2014real}: basic network (first row) and region ensemble network (second row).}
\label{fig_vis}
\end{figure*}
We apply our method on two publicly datasets: ICVL \cite{tang2014latent} and NYU \cite{tompson2014real}. The former dataset has 300K images for training and 1.6K for testing with 16 joints. The latter dataset has 72K images for training and 8K for testing with 14 joints. The performance is evaluated by two metrics: per-joint average Euclidean distance (in millimeters) and percentage of frames in which all errors of joints are below a threshold \cite{oberweger2015hands}. First we compare our REN with baseline and different ensemble settings on ICVL dataset. Then we compare it with several state-of-the-art methods on both datasets.

\noindent\textbf{Self-comparison}\hspace{2mm}
For comparison, we implement four baseline: 1) \emph{Basic} network has the same convolution structure in Fig.\ref{fig_baseline} and single regressor on the full feature map. 2) \emph{Basic Large} network is the same as basic network except for using 8192 dimensions in the second FC layer, which contains the similar number of parameters to REN. 3) \emph{Basic Bagging} network has four basic networks trained independently on the same data with different random order and augmentation. The average predictions of all the networks form the final prediction. 4) \emph{Region Bagging} network shares the same region division with REN but predicts independent hand pose for each region and averages them as prediction. 

\begin{figure}[htb]
\centering
{\includegraphics[width=0.5\textwidth]{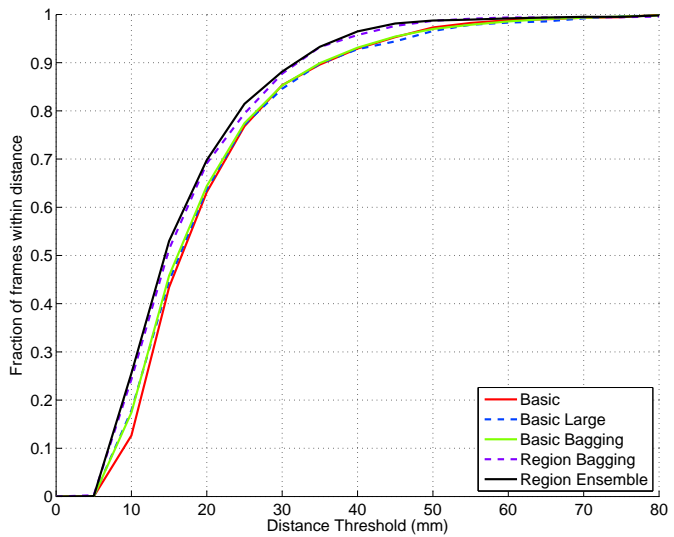}}
\caption{Self-comparison for percentage of success frames on ICVL dataset \cite{tang2014latent}. }
\label{fig_self}
\end{figure}

Results in Fig.\ref{fig_self} shows that: 1) bagging or ensemble based methods are more effective since all the ensemble versions significantly outperform the single one. 2) region ensemble is much better than basic bagging and slightly better than region bagging. Qualitative results of region ensemble (second row) and basic network (first row) are shown in Fig.\ref{fig_vis}. 

Table.1 further compares the running time (Nvidia Titan X GPU) for different approaches. Our REN obtains the most accurate results with nearly the same number of parameters as basic large network and region bagging, while the basic bagging takes significantly more parameters and time. And it runs up to over 3000fps on a single GPU, which is fast enough for practical use.

\begin{table}[htb]
\label{table_result}
\caption{Average 3D distance error (mm) and GPU forward time (ms) of different methods on ICVL dataset \cite{tang2014latent}.}
\centering
\begin{tabular}{|c|c|c|c|}
\hline
Method & Error(mm) & Time(ms)\\
\hline
Basic & 8.36 & 0.21\\
Basic Large & 8.18 & 0.22\\
Basic Bagging & 7.94 & 0.88\\
Region Bagging & 7.63 & 0.31\\
Region Ensemble & \textbf{7.47} & 0.31 \\
\hline
\end{tabular}
\end{table}

\begin{figure*}[htb]
\centering
\begin{minipage}[b]{0.49\textwidth}
  \centering
  \centerline{\includegraphics[width=0.95\textwidth]{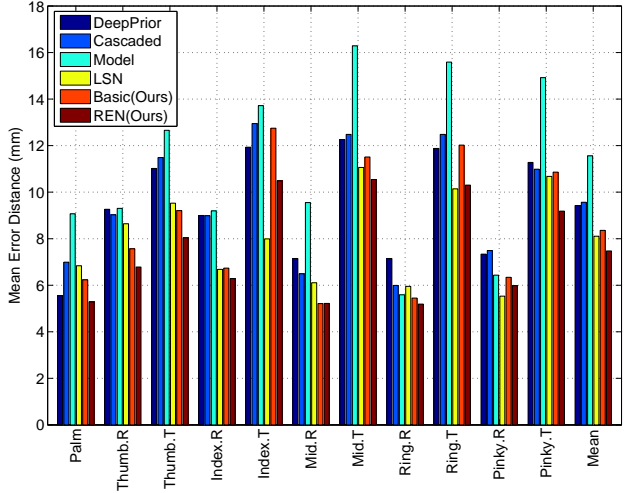}}
\end{minipage}
\begin{minipage}[b]{0.49\textwidth}
  \centering
  \centerline{\includegraphics[width=0.95\textwidth]{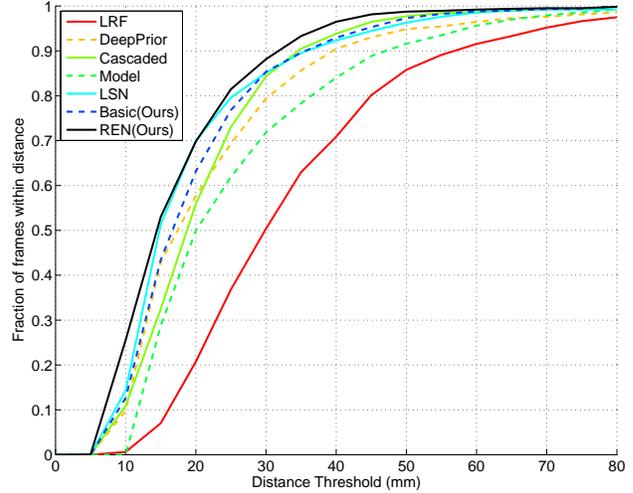}}
\end{minipage}
\caption{Comparison with state-of-the-arts on ICVL \cite{tang2014latent} dataset: distance error (left) and percentage of success frames (right).}
\label{fig_sota_icvl}
\end{figure*}

\begin{figure*}[htb]
\begin{minipage}[b]{0.49\textwidth}
  \centering
  \centerline{\includegraphics[width=0.95\textwidth]{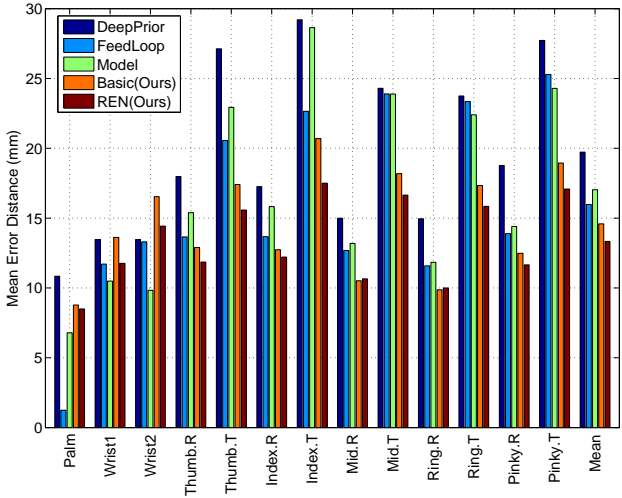}}
\end{minipage}
\begin{minipage}[b]{0.49\textwidth}
  \centering
  \centerline{\includegraphics[width=0.95\textwidth]{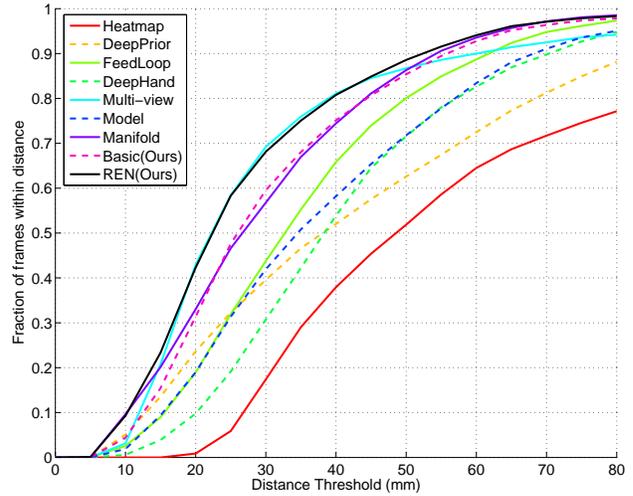}}
\end{minipage}
\caption{Comparison with state-of-the-arts on NYU \cite{tompson2014real} datasets: distance error (left) and percentage of success frames (right).}
\label{fig_sota_nyu}
\end{figure*}

\noindent\textbf{Comparison with state-of-the-arts}\hspace{2mm} We compare our methods against several state-of-the-art approaches on ICVL dataset \cite{tang2014latent} \cite{oberweger2015hands} \cite{sun2015cascaded} \cite{zhou2016model} \cite{wan2016hand} and NYU dataset \cite{tompson2014real} \cite{oberweger2015hands} \cite{oberwegertraining} \cite{sinha2016deephand} \cite{gerobust} \cite{zhou2016model} \cite{zhang2016learning}. Fig.\ref{fig_sota_icvl} and \ref{fig_sota_nyu} show that proposed REN obtains the best accuracy among all the algorithms.

In details, on ICVL our method outperforms LSN \cite{wan2016hand} on the threshold of $(5mm, 15mm)$ and $(20mm, 60mm)$, and surpasses other methods with a large margin. And the mean errors obtains $0.63mm$ decrease compared with LSN, which is a $7.77\%$ relative improvement. Similarly on NYU, our results are better than multi-view ConvNets \cite{gerobust} on the threshold of $(5mm, 15mm)$ and $(40mm, 80mm)$ and significantly more accurate than other approaches. Note that either LSN or multi-view ConvNets employ multiple models with complicated inputs, while our REN only uses single model without multi-stage regression, which indicates the power for proposed region ensemble strategy.

\section{Conclusion}
To boost the performance of single ConvNet for hand pose estimation, we present a simple but powerful region ensemble structure by dividing the feature maps into different regions and jointly training multiple regressors on all regions with fusion. Such strategy significantly improves the ConvNet without extra large computation overhead. The experimental results demonstrate that our method outperforms all the state-of-the-arts on two datasets. In the future we will investigate and analysis more ensemble methods for ConvNets. Since proposed region ensemble is general, we will also try to apply them on more tasks such as human pose estimation.

\noindent\textbf{Acknowledgments}\hspace{2mm} This work is supported by NSFC (No. 61271390), and 863 Plan (No. 2015AA016304). Thanks Shulan Pan for paper edition.

% References should be produced using the bibtex program from suitable
% BiBTeX files (here: strings, refs, manuals). The IEEEbib.bst bibliography
% style file from IEEE produces unsorted bibliography list.
% -------------------------------------------------------------------------
\bibliographystyle{IEEEbib}
\bibliography{refs}

\begin{thebibliography}{10}

\bibitem{supancic2015depth}
James~Steven Supancic~III, Gregory Rogez, Yi~Yang, Jamie Shotton, and Deva
  Ramanan,
\newblock ``Depth-based hand pose estimation: methods, data, and challenges,''
\newblock in {\em IEEE International Conference on Computer Vision}, 2015.

\bibitem{krizhevsky2012imagenet}
Alex Krizhevsky, Ilya Sutskever, and Geoffrey~E Hinton,
\newblock ``Imagenet classification with deep convolutional neural networks,''
\newblock in {\em Advances in neural information processing systems}, 2012, pp.
  1097--1105.

\bibitem{bulat2016human}
Adrian Bulat and Georgios Tzimiropoulos,
\newblock ``Human pose estimation via convolutional part heatmap regression,''
\newblock in {\em European Conference on Computer Vision}. Springer, 2016, pp.
  717--732.

\bibitem{tompson2014real}
Jonathan Tompson, Murphy Stein, Yann Lecun, and Ken Perlin,
\newblock ``Real-time continuous pose recovery of human hands using
  convolutional networks,''
\newblock {\em ACM Transactions on Graphics}, vol. 33, no. 5, pp. 169, 2014.

\bibitem{oberweger2015hands}
Markus Oberweger, Paul Wohlhart, and Vincent Lepetit,
\newblock ``Hands deep in deep learning for hand pose estimation,''
\newblock {\em Computer Vision Winter Workshop}, 2015.

\bibitem{oberwegertraining}
Markus Oberweger, Paul Wohlhart, and Vincent Lepetit,
\newblock ``Training a feedback loop for hand pose estimation,''
\newblock in {\em IEEE International Conference on Computer Vision}, 2015.

\bibitem{gerobust}
Liuhao Ge, Hui Liang, Junsong Yuan, and Daniel Thalmann,
\newblock ``Robust 3d hand pose estimation in single depth images: from
  single-view cnn to multi-view cnns,''
\newblock in {\em IEEE Conference on Computer Vision and Pattern Recognition},
  2016.

\bibitem{zhang2016learning}
Yu~Zhang, Chi Xu, and Li~Cheng,
\newblock ``Learning to search on manifolds for 3d pose estimation of
  articulated objects,''
\newblock {\em arXiv preprint arXiv:1612.00596}, 2016.

\bibitem{sun2015cascaded}
Xiao Sun, Yichen Wei, Shuang Liang, Xiaoou Tang, and Jian Sun,
\newblock ``Cascaded hand pose regression,''
\newblock in {\em IEEE Conference on Computer Vision and Pattern Recognition},
  2015, pp. 824--832.

\bibitem{wan2016hand}
Chengde Wan, Angela Yao, and Luc Van~Gool,
\newblock ``Hand pose estimation from local surface normals,''
\newblock in {\em European Conference on Computer Vision}, 2016.

\bibitem{tang2014latent}
Danhang Tang, Hyung~Jin Chang, Alykhan Tejani, and Tae-Kyun Kim,
\newblock ``Latent regression forest: Structured estimation of 3d articulated
  hand posture,''
\newblock in {\em IEEE Conference on Computer Vision and Pattern Recognition}.
  IEEE, 2014, pp. 3786--3793.

\bibitem{wang2013depth}
Guijin Wang, Xuanwu Yin, Xiaokang Pei, and Chenbo Shi,
\newblock ``Depth estimation for speckle projection system using progressive
  reliable points growing matching,''
\newblock {\em Applied optics}, vol. 52, no. 3, pp. 516--524, 2013.

\bibitem{shi2015high}
Chenbo Shi, Guijin Wang, Xuanwu Yin, Xiaokang Pei, Bei He, and Xinggang Lin,
\newblock ``High-accuracy stereo matching based on adaptive ground control
  points,''
\newblock {\em IEEE Transactions on Image Processing}, vol. 24, no. 4, pp.
  1412--1423, 2015.

\bibitem{zhou2016model}
Xingyi Zhou, Qingfu Wan, Wei Zhang, Xiangyang Xue, and Yichen Wei,
\newblock ``Model-based deep hand pose estimation,''
\newblock in {\em IJCAI}, 2016.

\bibitem{sun2013deep}
Yi~Sun, Xiaogang Wang, and Xiaoou Tang,
\newblock ``Deep convolutional network cascade for facial point detection,''
\newblock in {\em IEEE Conference on Computer Vision and Pattern Recognition},
  2013, pp. 3476--3483.

\bibitem{walach2016learning}
Elad Walach and Lior Wolf,
\newblock ``Learning to count with cnn boosting,''
\newblock in {\em European Conference on Computer Vision}. Springer, 2016, pp.
  660--676.

\bibitem{guo2016two}
Hengkai Guo, Guijin Wang, and Xinghao Chen,
\newblock ``Two-stream convolutional neural network for accurate rgb-d
  fingertip detection using depth and edge information,''
\newblock {\em arXiv preprint arXiv:1612.07978}, 2016.

\bibitem{li2016deeptrack}
Hanxi Li, Yi~Li, and Fatih Porikli,
\newblock ``Deeptrack: Learning discriminative feature representations online
  for robust visual tracking,''
\newblock {\em IEEE Transactions on Image Processing}, vol. 25, no. 4, pp.
  1834--1848, 2016.

\bibitem{chen2016accurate}
Xinghao Chen, Guijin Wang, and Hengkai Guo,
\newblock ``Accurate fingertip detection from binocular mask images,''
\newblock in {\em Visual Communications and Image Processing (VCIP), 2016}.
  IEEE, 2016, pp. 1--4.

\bibitem{li2016convolutional}
Hanxi Li, Yi~Li, and Fatih Porikli,
\newblock ``Convolutional neural net bagging for online visual tracking,''
\newblock {\em Computer Vision and Image Understanding}, pp. 120--129, 2016.

\bibitem{ahmed2016network}
Karim Ahmed, Mohammad~Haris Baig, and Lorenzo Torresani,
\newblock ``Network of experts for large-scale image categorization,''
\newblock {\em arXiv preprint arXiv:1604.06119}, 2016.

\bibitem{sermanet2013overfeat}
Pierre Sermanet, David Eigen, Xiang Zhang, Micha{\"e}l Mathieu, Rob Fergus, and
  Yann LeCun,
\newblock ``Overfeat: Integrated recognition, localization and detection using
  convolutional networks,''
\newblock {\em arXiv preprint arXiv:1312.6229}, 2013.

\bibitem{he2015deep}
Kaiming He, Xiangyu Zhang, Shaoqing Ren, and Jian Sun,
\newblock ``Deep residual learning for image recognition,''
\newblock {\em arXiv preprint arXiv:1512.03385}, 2015.

\bibitem{jia2014caffe}
Yangqing Jia, Evan Shelhamer, Jeff Donahue, Sergey Karayev, Jonathan Long, Ross
  Girshick, Sergio Guadarrama, and Trevor Darrell,
\newblock ``Caffe: Convolutional architecture for fast feature embedding,''
\newblock {\em arXiv preprint arXiv:1408.5093}, 2014.

\bibitem{sinha2016deephand}
Ayan Sinha, Chiho Choi, and Karthik Ramani,
\newblock ``Deephand: robust hand pose estimation by completing a matrix
  imputed with deep features,''
\newblock in {\em IEEE Conference on Computer Vision and Pattern Recognition},
  2016.

\end{thebibliography}

\end{document}